\documentclass[conference]{IEEEtran}
\IEEEoverridecommandlockouts
\usepackage{cite}
\usepackage{url}
\usepackage{booktabs}
\usepackage{amsmath,amssymb,amsfonts}
\usepackage{algorithmic}
\usepackage{graphicx}
\usepackage{lipsum}
\usepackage[T1]{fontenc}
\usepackage{textcomp}
\usepackage{tikz}
\usepackage{atbegshi}
\usepackage{xcolor}
\usepackage{csquotes}
\usepackage{comment}
\usepackage{hyperref}
 \hypersetup{
colorlinks=true,
linkcolor=blue,
filecolor=blue,
citecolor = blue,
urlcolor=blue,
}
\def\BibTeX{{\rm B\kern-.05em{\sc i\kern-.025em b}\kern-.08em
    T\kern-.1667em\lower.7ex\hbox{E}\kern-.125emX}}
\begin{document}

\title{Machine Learning Meets Transparency in Osteoporosis Risk Assessment: A Comparative Study of ML and Explainability Analysis}

\author{
\IEEEauthorblockN{Farhana Elias}
\IEEEauthorblockA{\textit{Dept. of ECE} \\
\textit{North South University}\\
Dhaka-1229, Bangladesh \\
farhana.elias@northsouth.edu}
\and
\IEEEauthorblockN{Md Shihab Reza}
\IEEEauthorblockA{\textit{Dept. of ECE} \\
\textit{North South University}\\
Dhaka-1229, Bangladesh \\
shihab.reza@northsouth.edu}
\and
\IEEEauthorblockN{Muhammad Zawad Mahmud}
\IEEEauthorblockA{\textit{Dept. of ECE} \\
\textit{North South University}\\
Dhaka-1229, Bangladesh \\
zawad.mahmud1@northsouth.edu}
\and
\IEEEauthorblockN{Samiha Islam}
\IEEEauthorblockA{\textit{Dept. of ECE} \\ 
\textit{North South University}\\
Dhaka-1229, Bangladesh \\
samiha.islam2@northsouth.edu}
\and
\IEEEauthorblockN{Shahran Rahman Alve}
\IEEEauthorblockA{\textit{Dept. of ECE} \\ 
\textit{North South University}\\
Dhaka-1229, Bangladesh \\
shahran.alve@northsouth.edu}
}

\maketitle

\begin{abstract}
The present research tackles the difficulty of predicting osteoporosis risk via machine learning (ML) approaches, emphasizing the use of explainable artificial intelligence (XAI) to improve model transparency. Osteoporosis is a significant public health concern, sometimes remaining untreated owing to its asymptomatic characteristics, and early identification is essential to avert fractures. The research assesses six machine learning classifiers—Random Forest, Logistic Regression, XGBoost, AdaBoost, LightGBM, and Gradient Boosting—utilizing a dataset based on clinical, demographic, and lifestyle variables. The models are refined using GridSearchCV to calibrate hyperparameters, with the objective of enhancing predictive efficacy. XGBoost had the greatest accuracy (91\%) among the evaluated models, surpassing others in precision (0.92), recall (0.91), and F1-score (0.90). The research further integrates XAI approaches, such as SHAP, LIME, and Permutation Feature Importance, to elucidate the decision-making process of the optimal model. The study indicates that age is the primary determinant in forecasting osteoporosis risk, followed by hormonal alterations and familial history.  These results corroborate clinical knowledge and affirm the models' therapeutic significance. The research underscores the significance of explainability in machine learning models for healthcare applications, guaranteeing that physicians can rely on the system's predictions. The report ultimately proposes directions for further research, such as validation across varied populations and the integration of supplementary biomarkers for enhanced predictive accuracy.
\end{abstract}


\section{Introduction}
 Osteoporosis (OP) is a common condition that leads to fragility fractures because of   a systemic decrease in bone mass and microarchitectures. Bone remodeling is a lifelong process that  involves  a continuous cycle of bone resorption and formation. One in three women and one in five  men over  the age of 50 are affected by osteoporosis worldwide [13]. The global population of 158 million people aged 50 and above faces high risk of  osteoporotic fractures and The risk of osteoporotic fractures is expected to double in 2045 due to  progressive ageing \cite{oden2015burden}. Recent studies indicate that approximately 37.3\% of Bangladeshi adults are affected by osteoporosis, with an additional 43.5\% experiencing osteopenia \cite{kha2023unveiling}. The lack of access to Dual-energy X-ray Absorptiometry (DXA) traditional  diagnostic modalities persists in rural and resource-constrained settings because of high costs and limited availability. The  situation requires new scalable methods for early detection and risk stratification. Machine Learning (ML) presents a  promising solution because it uses clinical data together with demographic information and lifestyle patterns to make accurate osteoporosis  risk predictions. ML models show promising results for predicting fracture risk in patients with  osteoporosis. The \enquote{black-box} nature of many ML algorithms poses challenges in clinical adoption, as healthcare professionals  require transparency to trust and act upon model predictions. This is where Explainable Artificial Intelligence  (XAI) becomes crucial. By elucidating the rationale behind predictions, XAI enhances interpretability, allowing clinicians  to understand and validate the factors contributing to an individual's risk assessment. Such transparency is vital for informed  decision-making and patient education. 
The motivation of this study is to design and evaluate multiple ML classifiers and reveal the influential features for the model decision making process through XAI. The main contributions of this research are as follows: 
\begin{itemize}
    \item Comparison of several ML classifiers such as RandomForest, Logistic Regression, XGBoost, AdaBoost, LightGBM and Gradient Boosting and integrating GridSearchCV to tune the best hyperparameters of the applied models.
    \item The application of the most explainable AI techniques (SHAP, LIME, and Permutation Feature Importance) to understand how the models work and the role of the features in making decisions.
\end{itemize}

The integration of ML with XAI in osteoporosis risk assessment systems will transform early detection strategies  in Bangladesh. These systems can enable targeted interventions, optimize resource allocation, and ultimately reduce the incidence of  osteoporotic fractures. The high prevalence and underdiagnosis of osteoporosis in the country makes  the deployment of ML and XAI-driven tools both timely and necessary in public health.

\section{Related Work}
Several researchers have conducted numerous ML studies on osteoarthritis prediction and risk assessment.

De Vries et al. \cite{de2021comparing} developed four machine learning models to forecast the probability of future major osteoporotic fractures (MOF).
In study \cite{jabarpour2020osteoporosis} collected personal, lifestyle, and clinical data from individuals and used Support Vector Machines (SVM) and Tree-Augmented Naïve Bayes (TAN) classifiers to mine patterns and predict disease status. 
In study \cite{yoo2013risk} researchers with the aim of identifying the risk of femoral neck osteoporosis in postmenopausal women developed an ML model and compared those to a conventional clinical decision tool, the osteoporosis self-assessment tool (OST). 
Mahmud et al. \cite{mahmud2024optimizing} Showed how feature selection techniques, such as Lasso L1, improve model accuracy and scalability in predictive systems.
Zeitlin et al. \cite{zeitlin2023clinical} developed an ML model for predicting 10-year risk of menopause-related osteoporosis by using clinical, laboratory, and imaging data, and findings revealed that the 10-year osteoporosis prediction model demonstrated strong discrimination (AUC = 0.83, Brier = 0.054), with spine and hip BMD plus age as the top predictors, and risk stratification thresholds yielding sensitivity 81\%, specificity 82\%, and likelihood ratios of 0.23 (low), 3.2 (medium), and 6.8 (high). 
A novel approach for the early detection of osteoporosis using ML and simulation tools based on the biomarker was presented in study \cite{budyal2024early}. Reza et al. \cite{reza2024linear} demonstrated how LDA as a feature reduction technique can be useful for reducing the burden of the model's complexity in predictive systems. 
Shim et al. \cite{shim2020application}, by using KNHANES V-1 and V-2 data from 1,792 postmenopausal Korean women, built seven ML models and found the ANN model achieved the highest AUROC (0.743) for osteoporosis risk prediction. S. K. Kim et al. [10] developed and validated ML models (SVM, RF, ANN, LR) using KNHANES-V1 data to predict osteoporosis risk in postmenopausal Korean women.    

Advanced DL and XAI techniques can also be utilized while predicting osteoporosis or its risk assessment. 
Qiu et al. \cite{qiu2024developing} investigated whether DNN can achieve a better performance in osteoporosis risk prediction, and their study revealed that a deep neural network using 16 routine clinical and demographic variables achieved the highest osteoporosis risk prediction performance. 
Mahmud et al. \cite{mahmud2024advance} also demonstrated how effective transfer learning approach can be in disease predictive systems. 
In study \cite{khanna2023decision}, researchers developed an ML pipeline by the forward feature selection and a custom multi-level ensemble stack and achieved 89\% prediction accuracy and used SHAP, LIME, ELI5, and Qlattice for model explainability.

\section{Methodology}
\begin{figure}[h]
    \centering
    \includegraphics[width=0.45\textwidth]{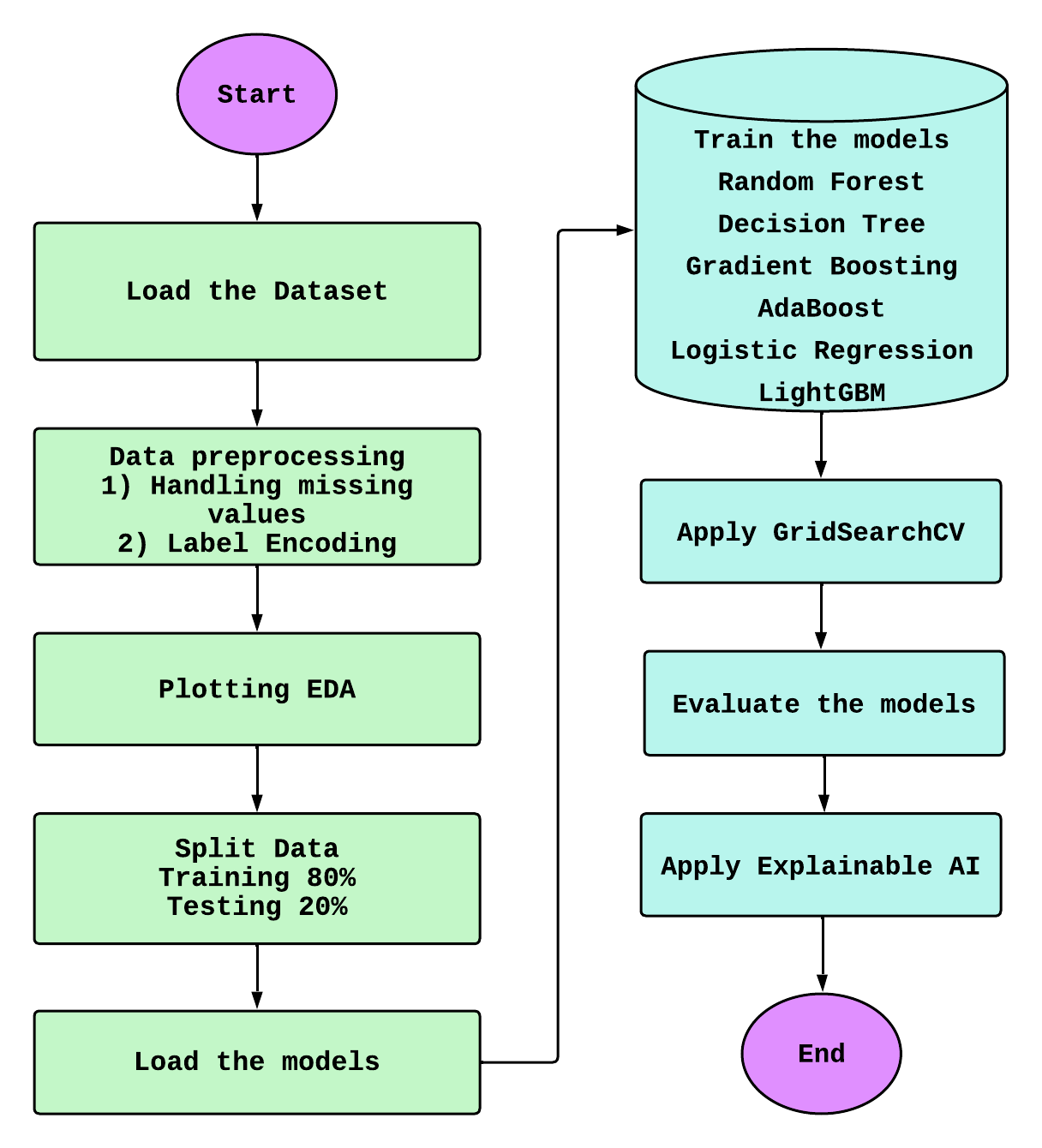}
    \caption {Workflow diagram of the system}
    \label{fig:1}
\end{figure}

This section outlines the methodological approach implemented in the study. Fig. \ref{fig:1} illustrates the corresponding workflow diagram, which summarizes the key stages of the research process.

\subsection{Dataset}
The data for this paper named \enquote*{Osteoporosis Risk Prediction} was obtained from Kaggle \cite{Kulkarni2024}. It comprises 1,958 entries detailing various health and lifestyle attributes pertinent to  osteoporosis risk assessment. Key features include demographic details (age, gender, race/ethnicity),  lifestyle habits (physical activity, smoking, alcohol consumption), dietary intakes (calcium, vitamin  D), and medical history (hormonal changes, family history of osteoporosis). We divided the dataset into two parts: training set 80\% and test set 20\% to try out different ML models.

\subsection{Exploratory Data Analysis (EDA)}
\begin{figure}[h]
    \centering
    \includegraphics[width=0.45\textwidth]{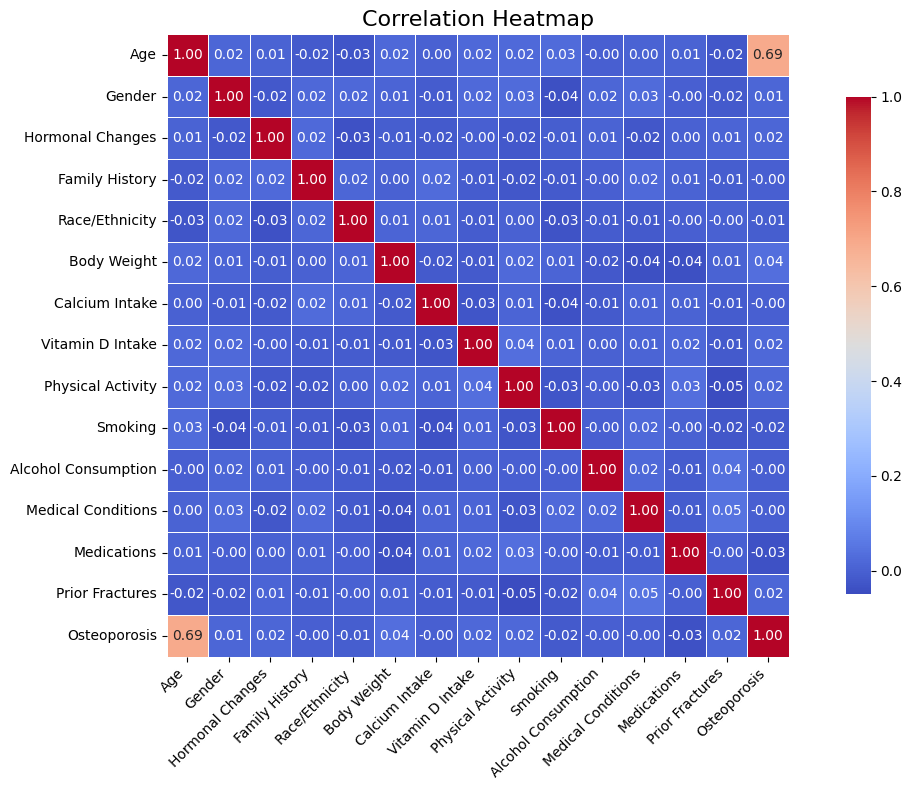}
    \caption {Correlation matrix}
    \label{fig:2}
\end{figure}

Several exploratory analyses were conducted to assess relationships between lifestyle/medical variables and osteoporosis diagnosis. The data revealed no stark imbalances in most variables, suggesting weak or non-linear associations in this dataset. For instance, both hormonal changes (e.g., 498 non-osteoporosis vs. 483 osteoporosis cases without hormonal changes) and family history (498 vs. 500 cases without family history) showed near-even splits, implying minimal direct predictive power. Similarly, race/ethnicity and medical conditions categories displayed uniform distributions across osteoporosis outcomes, with no subgroup exceeding 344 cases. Body weight demonstrated a minimal pattern where people with lower body weight (0) exhibited slightly elevated  osteoporosis rates (531 non-osteoporosis vs. 496 osteoporosis) than those  with higher body weight (1). The nutritional factors calcium intake (475 vs. 479  low-intake osteoporosis cases) and vitamin D intake (482 vs. 465 low-intake  cases) demonstrated minimal variations but vitamin D intake above the threshold resulted in a minimal rise of osteoporosis  diagnosis (497 vs. 514).   The distribution of lifestyle habits including physical activity (520 vs. 501 low-activity  osteoporosis cases), smoking (480 vs. 496 non-smokers with osteoporosis), and alcohol  consumption (484 vs. 486 abstainers with osteoporosis) showed balanced distributions. The distribution  of prior fractures and medications followed the same pattern as other variables since osteoporosis cases showed less than  20 counts between categories. The results suggest minor patterns (e.g., medications show a slight decrease  in osteoporosis risk between 509 and 476 cases). Fig. \ref{fig:2} presents the correlation matrix of the dataset variables, highlighting interdependencies between features. 
\subsection{Models Applied and Evaluation}
In this study, we applied 6 different ML classifiers: RandomForest, Logistic Regression, XGBoost, AdaBoost, LightGBM and Gradient Boosting with a 5-fold cross-validation-integrated Grid- SearchCV hyperparameter tuning framework to  improve its performance. Random forests are an ensemble learning method introduced by Breiman et al. \cite{breiman2001random}.  Logistic regression \cite{cox1958regression} is a supervised machine learning algorithm used for binary classification tasks, modelling the probability of a binary outcome based on one or more predictor variables. XGBoost (eXtreme Gradient Boosting) is a scalable machine learning system for tree boosting, developed by Chen et al. \cite{chen2016xgboost}. AdaBoost (Adaptive Boosting) \cite{yulianto2019improving} sequentially trains weak classifiers (typically decision-tree stumps) by increasing the weights of previously misclassified samples so that new learners concentrate on the hardest cases.  LightGBM (Light Gradient-Boosting Machine) \cite{zhang2019lightgbm} is an open-source, highly efficient gradient-boosting framework. Gradient Boosting 
constructs an ensemble by fitting each new weak learner to the negative gradient (residuals) of a specified loss function with respect to the current model’s predictions. 
For result evaluation, we used five state-of-the-art metrics for osteoporosis risk scoring: accuracy, precision, recall, and F1-score. The metrics which were used to evaluate the models are shown below \cite{islam2024deep,mahmud2024op}: 

\begin{equation}
\label{eq:precision}
\text{Precision}_x = \frac{TP_x}{TP_x + FP_x}
\end{equation}

\begin{equation}
\label{eq:recall}
\text{Recall}_x = \frac{TP_x}{TP_x + FN_x}
\end{equation}

\begin{equation}
\label{eq:F1}
F1-score_x= 2 \times \frac{\text{Precision}_x \times \text{Recall}_x}{\text{Precision}_x + \text{Recall}_x}
\end{equation}

\begin{equation}
\label{eq:accuracy}
\text{Accuracy}_x = \frac{TP_x + TN_x}{TP_x + TN_x + FP_x + FN_x}
\end{equation}
\subsection{Explainability of Model}
Several XAI techniques such  SHAP, LIME, Morris Sensitivity Analysis, and Permutation Feature Importance to understand how the models work and the role of the features in making decisions. SHAP (SHapley Additive exPlanations) is a unified framework for interpreting model predictions, as introduced by Lundberg et al. [23].  Local Interpretations of Model-Agnostic Explanations (LIME) by Ribeiro et al. [24] is a XAI method used to produce human-interpretable and instance-specific explanations for complex ML models. One of the global explainable AI (XAI) techniques that determines the significance of a model is Permutation Feature Importance (PFI) [25].  Integrating these XAI techniques not only reveals the most influential features but also ensures transparency in the model’s decision-making process, making the model more reliable and trustworthy. 

\section{Result Analysis and Discussion}
In this section, we have evaluated and compared the performance of all ML models implemented  in this study. Explainable AI tools were deployed for the best-performing  model, along with a comprehensive understanding of the feature importance. 

\subsection{Machine Learning Models For Osteoporosis Risk Prediction}

Table \ref{tab:1} summarizes the optimized performance of six machine-learning models on osteoporosis risk prediction. XGBoost achieved the highest accuracy at 91.0\%, outperforming all other methods; it also led on precision (0.92), recall (0.91), and F1-score (0.90), indicating both strong overall classification and balanced sensitivity/specificity. LightGBM followed closely with 90.05\% accuracy and a 0.91/0.90 precision-recall pair, demonstrating that its leaf-wise growth and sampling optimizations yield nearly state-of-the-art results on this tabular health dataset. AdaBoost and Gradient Boosting produced comparable accuracies (89.0\%), with AdaBoost slightly edging in F1-score (0.88 vs. 0.89) and recall, suggesting that adaptive reweighting of hard examples remains highly effective under class imbalance. Random Forest and Logistic Regression, while simpler, still delivered respectable accuracies of 84.0\% and 83.67\%, respectively, confirming their baseline utility but highlighting the advantage of boosted trees for capturing complex feature interactions.

\begin{table}[htbp]
 \scriptsize
\caption{Performance Metrics Comparison}
\centering
\begin{tabular}{|c|c|c|c|c|}
\hline
\textbf{Model} & \textbf{Accuracy (\%)} &\textbf{Precision} & \textbf{Recall}&\textbf{F1 Score} \\
\hline
Random Forest (RF)
&84.00 
&0.88
&0.84
&0.84\\
\hline
Logistic Regression (LR)
&83.67
&0.88 
&0.84
&0.83 \\
\hline
XGBoost(XGB)
&91.0
&0.92
&0.91
&0.90\\
\hline
AdaBoost (AB)
&89.00
&0.90
&0.89  
&0.88\\
\hline
LightGBM (LGBM)
&90.05
&0.91 
&0.90
&0.90\\
\hline
Gradient Boosting (GB)
&89.0
&0.89 
&0.89 
&0.89 \\
\hline
\end{tabular}
\label{tab:1}
\end{table}

\begin{figure}[h]
    \centering
    \includegraphics[width=0.45\textwidth]{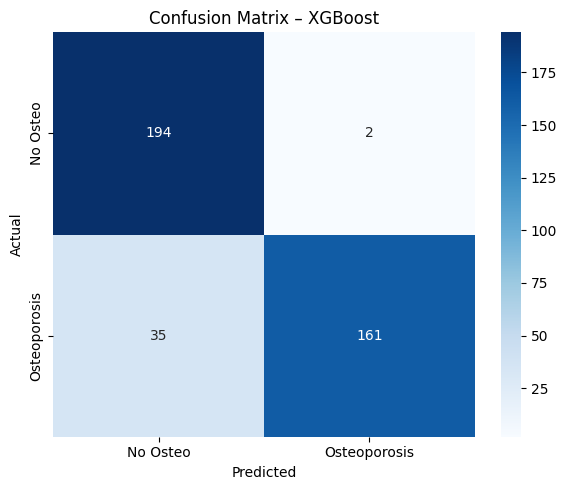}
    \caption{Confusion Matrix of XGBoost Model.}
    \label{fig:3}
\end{figure}

\begin{figure}[h]
    \centering
    \includegraphics[width=0.45\textwidth]{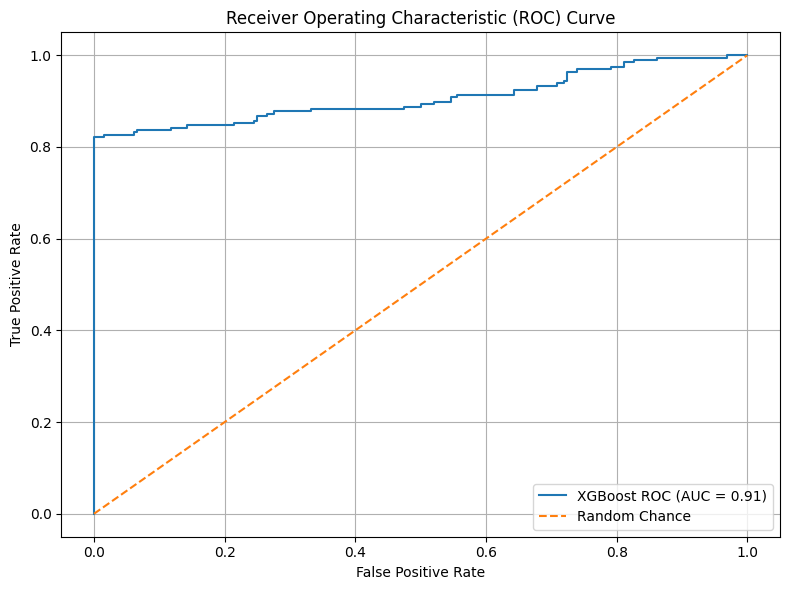}
    \caption{ROC curve of XGBoost Model.}
    \label{fig:4}
\end{figure}

Best Hyperparameters of the Applied ML models is shown in  Table \ref{tab:2}. These tuned hyperparameters reveal that moderate learning rates, regularization, and controlled tree complexity are key to maximizing predictive performance on osteoporosis risk data. The superior performance of boosted methods underscores their ability to model complex, non-linear relationships among demographic, clinical, and lifestyle factors. Fig. \ref{fig:3} shows the confusion matrix and Fig. \ref{fig:4} shows the ROC curve of the XGBoost model.

\begin{table}[htbp]
\tiny
\caption{Best Hyperparameters for Different Models}
\centering
\begin{tabular}{|c|c|}
\hline
\textbf{Model} & \textbf{Best Hyperparameters} \\
\hline
RF & \enquote*{criterion}: \enquote*{gini}, \enquote*{max\_depth}: 4, \enquote*{max\_features}: \enquote*{sqrt}, \enquote*{n\_estimators}: 100 \\
\hline
LR & \enquote*{C}: 0.01, \enquote*{penalty}: \enquote*{l1}, \enquote*{solver}: \enquote*{liblinear}  \\
\hline
XGB & \enquote*{max\_depth}: 3, \enquote*{reg\_lambda}: 0, \enquote*{min\_child\_weight}: 1, \enquote*{reg\_alpha}: 1.  \\
\hline
AB & \enquote*{algorithm}: \enquote*{SAMME}, \enquote*{learning\_rate}: 1, \enquote*{n\_estimators}: 50. \\
\hline
LGBM & \enquote*{learning\_rate}: 0.1, \enquote*{max\_depth}: 10, \enquote*{colsample\_bytree}: 0.8, \enquote*{n\_estimators}: 100.\\
\hline
GB & \enquote*{learning\_rate}: 0.1, \enquote*{max\_depth}: 8, \enquote*{n\_estimators}: 500, \enquote*{subsample}: 1. \\
\hline
\end{tabular}
\label{tab:2}
\end{table}

\subsection{Explainable artificial intelligence (XAI) to interpret XGBoost classifier predictions}
\begin{figure}[h]
    \centering
    \includegraphics[width=0.45\textwidth]{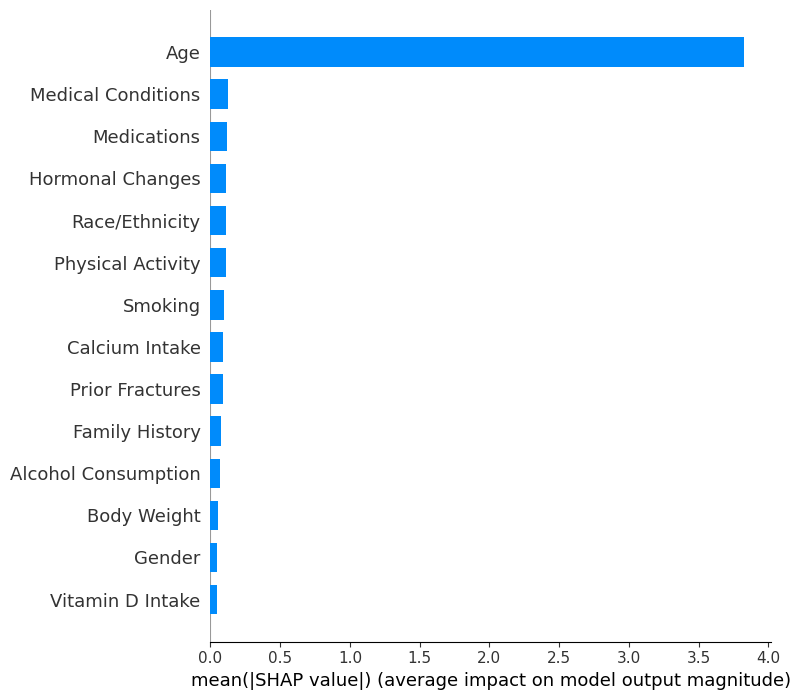}
    \caption{SHAP summary plot for the XGBoost model}
    \label{fig:5}
\end{figure}

\begin{figure}[h]
    \centering
    \includegraphics[width=0.45\textwidth]{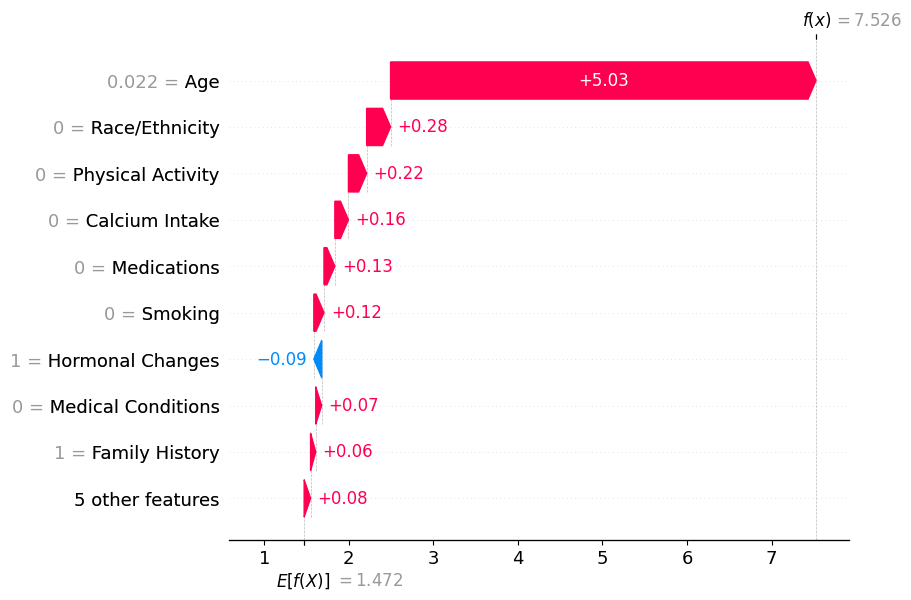}
    \caption{SHAP waterfall plot for the XGBoost model}
    \label{fig:6}
\end{figure}

In this section, we have evaluated and compared the influential features of the XGBoost classifiers. Explainable AI tools named SHAP, LIME, and permutation importance were deployed for the best-performing model, along with a comprehensive understanding of the feature importance. Fig. \ref{fig:5} and \ref{fig:6} show the SHAP summary plot and SHAP waterfall plot for the XGBoost model. 

The SHAP summary plot displays feature importance through mean absolute SHAP values, which show the average effect of each variable on model output deviation from the baseline prediction. Age stands as the most influential predictor because its mean SHAP value exceeds all other features by a significant amount. The medical field supports this observation because bone density naturally decreases with age progression. The model output receives its second-largest influence from medical conditions, medications, hormonal changes, race/ethnicity and physical activity. The remaining variables, including smoking, calcium intake, prior fractures, family history, alcohol consumption, body weight, gender and vitamin D intake, have small average effects on the model output. To unpack how these features combined to produce an individual prediction of 7.526 (baseline = 1.472), we examined the SHAP waterfall plot. Here, age again dominates, contributing +5.03 to the final risk score. Subtler positive effects arise from race/ethnicity (+0.28), physical activity (+0.22), calcium intake (+0.16), medications (+0.13) and smoking (+0.12). Notably, hormonal changes decrease the predicted risk by -0.09, suggesting a potential protective influence in this case. Smaller upward shifts stem from medical conditions (+0.07), family history (+0.06) and a combined boost of +0.08 from five other features.

\begin{figure}[h]
    \centering
    \includegraphics[width=0.45\textwidth]{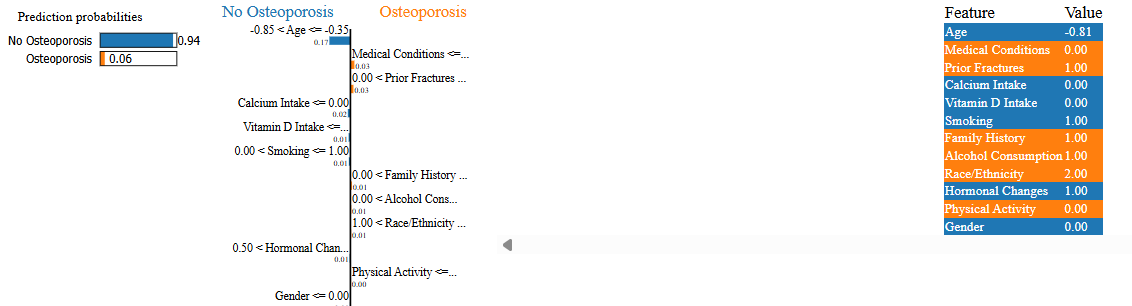}
    \caption{LIME feature-importance explanations for a randomly selected test-set sample using Model XGBoost.}
    \label{fig:7}
\end{figure}

The 0th test instance  receives a LIME explanation from the XGBoost model which predicts  \enquote{No Osteoporosis} with 94\% accuracy is represented in Fig. \ref{fig:7}. The prediction strongly points toward \enquote{No  Osteoporosis} because of three main factors: a younger age (-0.81  impact), male gender and adequate calcium and vitamin D intake. The other factors including medical conditions, prior  fractures, smoking, family history, alcohol use, race/ethnicity, hormonal changes and physical activity have  no significant impact.

\begin{figure}[h]
    \centering
    \includegraphics[width=0.45\textwidth]{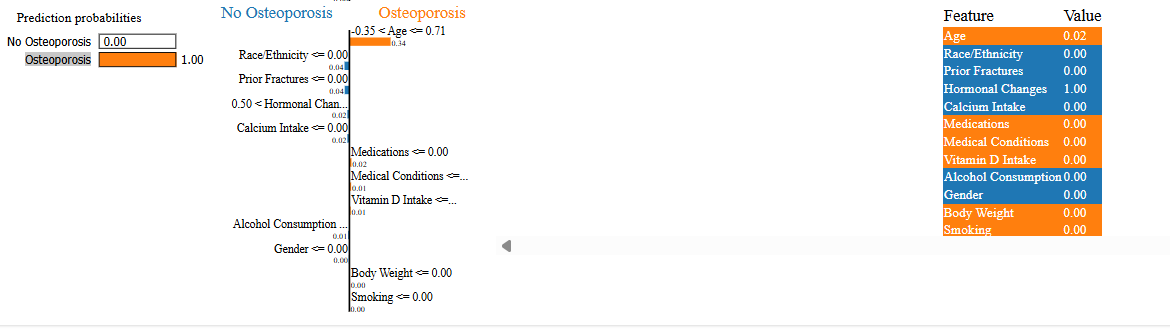}
    \caption{LIME feature-importance explanations for a different randomly selected test-set sample using Model XGBoost.}
    \label{fig:8}
\end{figure}

The LIME explanation presented in Fig. \ref{fig:8} shows the XGBoost model predicting the  100th test instance as \enquote{Osteoporosis} with absolute confidence of 100\%. The patient’s  age (0.02 on the normalized scale) stands as the main risk factor because it exists in  a higher risk range and generates the largest positive shift (+0.34) toward  \enquote{Osteoporosis.} The patient’s hormonal changes (value 1) slightly contribute to disease development while the remaining  features including race/ethnicity and prior fractures and calcium and vitamin D intake and medications and medical conditions  and alcohol use and gender and body weight and smoking have minimal or no influence. The combination of these  factors demonstrates why the model predicts this patient has osteoporosis. 

\begin{figure}[h]
    \centering
    \includegraphics[width=0.45\textwidth]{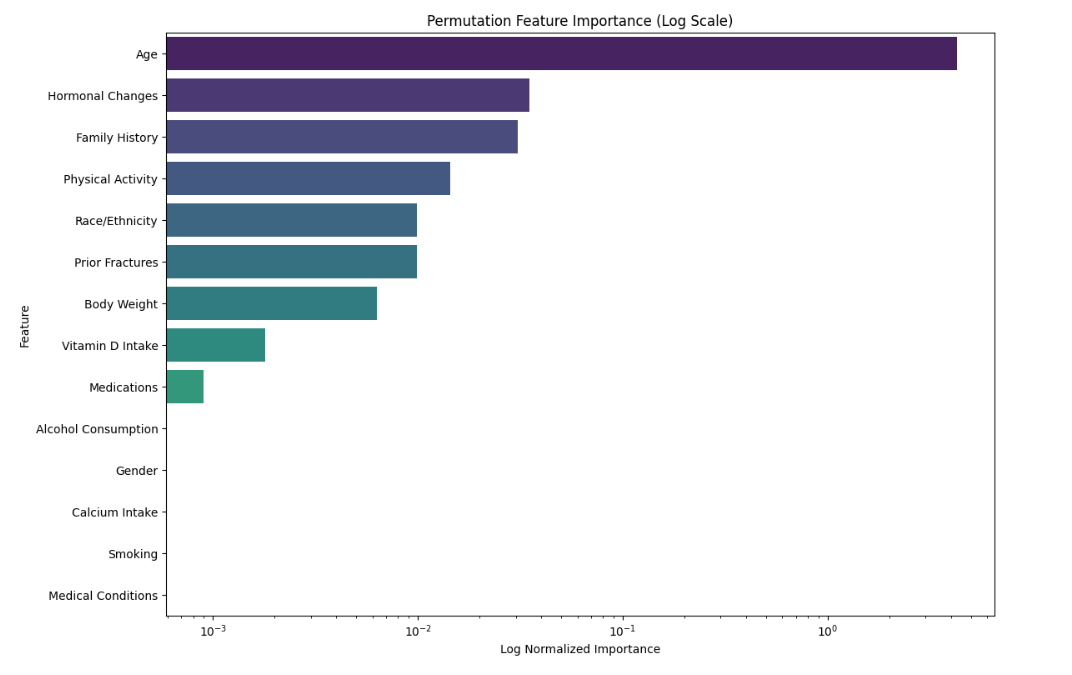}
    \caption{Permutation-importance rankings of features for a randomly selected test-set sample using Model XGBoost.}
    \label{fig:9}
\end{figure}

The permutation importance plot (log-scaled) in Fig. \ref{fig:9} shows that Age is by far the most influential predictor of osteoporosis risk; its importance is orders of magnitude higher than any other feature. The next most important factors are Hormonal Changes and Family History, which both contribute substantially but still far less than age.  Physical Activity, Race/Ethnicity, and Prior Fractures are moderately important. Body  Weight and Vitamin D Intake have smaller effects. Medications, Alcohol Consumption, Gender,  Calcium Intake, Smoking, and Medical Conditions are at the bottom, indicating they have little effect on  the model’s predictions once the other variables are controlled for.

The XGBoost model shows age as the primary risk factor for osteoporosis through all three XAI methods, which demonstrate its dominant effect over other variables. The XGBoost model shows age as the primary risk factor for osteoporosis through all three XAI methods, which demonstrate its dominant effect over other variables. The SHAP method assigns Hormonal Changes and Family History the second–third highest mean contributions, while permutation importance places them directly below age, and LIME shows their ability to affect individual predictions when present. Physical activity, race/ethnicity, and prior fractures form a middling group whose moderate importance appears across SHAP and permutation analyses and occasionally shapes LIME explanations. Finally, features such as calcium/vitamin D intake, medications, smoking, alcohol consumption, gender, body weight, and medical conditions consistently register minimal effect in all XAI views. This convergence of SHAP, LIME, and permutation importance gives us high confidence in the model’s reliance on age, hormonal status, and family history while indicating which other variables play only supporting or negligible roles.

\section{Conclusion and Future Work}
Osteoporosis is a pervasive public health challenge: initially asymptomatic, it ultimately reveals itself through debilitating fractures as bone strength deteriorates. The research evaluated six machine learning algorithms through systematic assessment for osteoporosis risk prediction tasks, including Random Forest, Logistic Regression, XGBoost, AdaBoost, LightGBM and Gradient Boosting. The evaluation of boosted tree methods through 5-fold cross-validated GridSearchCV revealed that these methods performed better than both bagged trees and linear classifiers. XGBoost demonstrated the highest performance among the models, with 91.0\% accuracy, 0.92 precision, 0.91 recall and 0.90 F1-score. Our analysis of hyperparameters revealed that balanced ensemble performance requires moderate tree depth settings together with learning rates around 0.1 and suitable regularisation and controlled sampling of features and data. The explainable AI methods SHAP, LIME and permutation importance revealed that age stands as the primary risk factor, while hormonal changes and family history rank as the second most important risk factors. 

The research contains certain limitations because the data distribution may differ from real-world patterns and lacks external validation from an independent cohort and does not include biomarkers or imaging or genetic data and cannot forecast when fractures will occur. Future research will investigate multiple directions. Our research will validate our model across different geographic regions through independent cohort studies to establish its applicability. Our research will expand feature collection by adding more biomarkers and imaging-derived metrics and genetic profiles to identify new predictive factors. Our research will advance past risk classification by developing survival analysis frameworks or recurrent architectures to predict both fracture timing and risk status. 
The data collection in Bangladesh will help create an XAI-based decision-support system which suits the needs of regional clinical practice. 

\bibliographystyle{IEEEtran}
\bibliography{IEEE.bib}

\end{document}